\documentclass[runningheads]{llncs}
\usepackage[british]{babel}
\usepackage[T1]{fontenc}

\usepackage{url}

\usepackage{mathtools} 
\usepackage{booktabs} 
\usepackage{longtable}
\usepackage{tikz} 

\usepackage[font=footnotesize]{subfig}
\usepackage{etoolbox}
\robustify\subref

\usepackage{xparse}
\DeclareDocumentCommand\datasetheading{ m g }{%
    \subsection*{#1%
        \IfNoValueF {#2} { \hfill \cite{#2}}%
    }%
}

\newcommand{\e}[1]{\emph{#1}}

\DeclareRobustCommand{\VAN}[3]{#2}

\begin{document}
\title{No Imputation without Representation}
%
%
\author{Oliver Urs Lenz\inst{1,2}\orcidID{0000-0001-9925-9482} \and
Daniel Peralta\inst{3}\orcidID{0000-0002-7544-8411} \and
Chris Cornelis\inst{2}\orcidID{0000-0002-6852-4041}}
\authorrunning{O. U. Lenz et al.}
%
\institute{Leiden Institute of Advanced Computer Science, Leiden University \email{o.u.lenz@liacs.leidenuniv.nl} \and
Research Group for Computational Web Intelligence,\\Department of Applied Mathematics, Computer Science and Statistics,\\Ghent University\\ \email{chris.cornelis@ugent.be} \url{https://cwi.ugent.be} \and
Department of Information Technology, Ghent University\\ \email{daniel.peralta@ugent.be}}
\maketitle              
\begin{abstract}
By filling in missing values in datasets, imputation allows these datasets to be used with algorithms that cannot handle missing values by themselves. However, missing values may in principle contribute useful information that is lost through imputation. The missing-indicator approach can be used in combination with imputation to instead represent this information as a part of the dataset. There are several theoretical considerations why missing-indicators may or may not be beneficial, but there has not been any large-scale practical experiment on real-life datasets to test this question for machine learning predictions. We perform this experiment for three imputation strategies and a range of different classification algorithms, on the basis of twenty real-life datasets. In a follow-up experiment, we determine attribute-specific missingness thresholds for each classifier above which missing-indicators are more likely than not to increase classification performance. And in a second follow-up experiment, we evaluate numerical imputation of one-hot encoded categorical attributes. We reach the following conclusions. Firstly, missing-indicators generally increase classification performance. Secondly, with missing-indicators, nearest neighbour and iterative imputation do not lead to better performance than simple mean/mode imputation. Thirdly, for decision trees, pruning is necessary to prevent overfitting. Fourthly, the thresholds above which missing-indicators are more likely than not to improve performance are lower for categorical attributes than for numerical attributes. Lastly, mean imputation of numerical attributes preserves some of the information from missing values. Consequently, when not using missing-indicators it can be advantageous to apply mean imputation to one-hot encoded categorical attributes instead of mode imputation.

\keywords{Missing data \and Missing-indicators \and Imputation \and Classification \and Data-centric machine learning.}
\end{abstract}

\section{Introduction}
\label{sec_introduction}

Missing values are a frequent issue in real-life datasets, and the subject of a large body of ongoing research. Some implementations of machine learning algorithms can handle missing values natively, requiring no further action by practitioners. But whenever this is not the case, a common general strategy is to replace the missing value with an estimated value: imputation. An advantage of imputation is that we obtain a complete dataset, to which we can apply any and all algorithms that make no special provision for missing values. However, missing values may be informative, and a disadvantage of imputation is that it removes this information.

The missing-indicator approach \cite{cohen68multiple} is an old proposal to represent and thereby preserve the information encoded by missing values. For every original attribute, it adds a new binary `indicator' or `dummy' attribute that takes a value of 1 if the value for the original attribute is missing, and 0 if not (Figure~\ref{fig_missing_indicators}).\footnote{Some authors use the opposite convention, letting the indicator express non-missingness.} The missing-indicator approach is often presented as an alternative to imputation, but since it does not resolve the missing values in the original attributes, it can only be used in addition to, not instead of imputation.

\begin{figure}
\centering
\begin{tabular}{p{.07\linewidth}p{.07\linewidth}p{.07\linewidth}p{.07\linewidth}p{.07\linewidth}p{.07\linewidth}}
$a_1$ & $a_2$ & $a_3$ & $i_1$ & $i_2$ & $i_3$\\
\midrule
0.92  & aap   & 2.50  & 0 & 0 & 0\\
?     & aap   & 1.00  & 1 & 0 & 0\\
8.42  & ?     & 3.00  & 0 & 1 & 0\\
2.23  & noot  & 0.05  & 0 & 0 & 0\\
?     & ?     & ?     & 1 & 1 & 1\\
0.41  & mies  & ?     & 0 & 0 & 1\\
$\vdots$ & $\vdots$ & $\vdots$ & $\vdots$ & $\vdots$ & $\vdots$\\
\end{tabular}%
\caption{Illustrative example of a dataset with three attributes ($a_1$, $a_2$, $a_3$) with missing values (`?'), and corresponding missing-indicators ($i_1$, $i_2$, $i_3$).}
\label{fig_missing_indicators}
\end{figure}

It is an open question whether missing-indicators should be used for predictive tasks in machine learning \cite{sperrin20missing}. Both imputation and the missing-indicator approach originate in the statistical literature. While imputation strategies have been the subject of a rich body of research, the missing-indicator approach has not received a large amount of attention, and is often dismissed or disregarded in overviews of approaches towards missing values.

In the context of machine learning, the effect of missing-indicators can be framed as follows. On the one hand, the addition of missing-indicators results in a more complete, higher-dimensional representation of the data. On the other hand, their omission corresponds to a form of dimensionality reduction, which may increase the efficiency and effectiveness of a dataset by eliminating redundancy.

To determine whether this trade-off is useful, a key question is to which extent missing values in a given data\-set are informative. If they are not, the phrase ``missing at random'' (MAR) \cite{rubin76inference} is used to indicate that the distribution of missing values is dependent on the known values, while the stricter phrase ``missing completely at random'' (MCAR) denotes values that are distributed truly randomly. In contrast, informative missing values are often denoted as ``missing not at random'' (MNAR).

In this respect, it is often argued that one should distinguish between missing values that could in principle have been obtained, and missing values that fundamentally do not exist, like attributes related to pregnancy tests for male subjects.\footnote{We are grateful to an anonymous reviewer for this example.} In the latter case, the missing values are definitely informative. However, such clear-cut cases may be comparatively rare. Moreover, it does not follow that the missing values in the former case are definitely non-informative. In fact, for real-life datasets, unless we have specific knowledge about the process responsible for the missing values, we have to assume some degree of informativeness in principle.\footnote{This is acknowledged by authors working under the assumption of MAR, e.g. ``When data are missing for reasons beyond the investigator's control, one can never be certain whether MAR holds. The MAR hypothesis in such datasets cannot be formally tested unless the missing values, or at least a sample of them, are available from an external source.'' \cite{schafer97analysis}}

Nonetheless, it has been argued that in practice, the attributes of a dataset can be sufficiently redundant that one can get away with assuming that its missing values are MAR \cite{schafer97analysis}. This means that most of the information contained by the missing values should in principle be recoverable through imputation. But even if this is so, imputation may not always perform optimally, in which case missing-indicators may still prove useful for machine learning.

A more subtle point is that even when missing values are informative, the information they encode need not be lost completely through imputation. This is particularly evident in the case of numerically encoded binary attributes (e.g. 0 and 1), where imputation can represent missing values as a third, intermediary value (e.g. 0.5). More generally, Le Morvan et al. \cite{lemorvan21whats} have recently observed that almost all deterministic imputation functions map records with missing values to distinct manifolds in the attribute space that can in principle be identified by sufficiently powerful algorithms. Nevertheless, missing-indicators can potentially make this learning task easier.

In light of these conflicting theoretical arguments, the usefulness of missing-indicators for real-life machine learning problems is an interesting empirical question. However, previous experiments in this direction have been limited in scope and number. These limitations include the use of only one or a handful of datasets, the use of datasets from which values have been removed artificially, and not comparing the same imputation strategies with and without missing-indicators.

The purpose of the present paper is straightforward. On the basis of twenty real-life classification problems with naturally occurring missing values, we measure the performance of a range of popular classification algorithms, using three common types of imputation, with and without missing-indicators. This allows us to evaluate the effect of using missing-indicators, as well as the choice of imputation strategy.

Moreover, we conduct three follow-up experiments to gain a better understanding of when and why missing-indicators can be useful. In the first, we determine whether this is influenced by the type (categorical or numerical) and the amount of missing values of a given attribute. In the second, we test the hypothesis that numerical imputation partially preserves the information from missing values. And in the third follow-up experiment, we compare missing-indicators to two model-specific approaches to missing values for nearest neighbour and decision tree classifiers.

In Section~\ref{sec_background}, we provide a brief overview of the existing literature on missing-indicators, including previous experimental evaluations. In Section~\ref{sec_experimental_setup}, we describe our experimental setup. We report our results in Section~\ref{sec_results} and conclude in Section~\ref{sec_conclusion}.
\section{Background}
\label{sec_background}

We start with a brief discussion of the origins and reception of the missing-indicator approach, as well as previous experimental evaluations of the use of missing-indicators in prediction tasks.

\subsection{Origins and Reception}
\label{sec_literature}

The missing-indicator approach originates in the literature on linear regression. It dates back to at least Cohen \cite{cohen68multiple}, who pointed out that values in real-life datasets are typically not missing completely at random, and that the distribution of missing values may in particular depend on the values of the attribute that is to be predicted. He proposed that each attribute could be said to have two `aspects', its value, and whether that value is present to begin with, which should be encoded with a pair of variables. For missing attribute values, the first of these variables was to be filled in with the mean of the known values, although other applications might call for different values. Cohen's proposal was subsequently expanded in \cite{cohen75applied}, but received only limited recognition in the following years \cite{kim77treatment,stumpf78note,chow79look,hutcheson81interpreting,anderson83missing,orme91multiple}.

Cohen's proposal was subjected to a formal analysis by Jones \cite{jones96indicator}, who showed that, if one assumes that missing values are MAR, and the true linear regression model does not contain any terms related to missingness, it produces biased estimates of the regression coefficients (unless the sample covariance between independent variables is zero). However, these assumptions run directly counter to the position set out in \cite{cohen75applied} that a priori, the missingness of each attribute is a possible explanatory factor, that it is safer not to assume that missing values are distributed randomly, and that the usefulness of missing-indicators is ultimately an empirical question.

Allison \cite{allison01missing}, motivated by \cite{jones96indicator} and working under the general assumption of MAR, dismissed missing-indicators as ``clearly unacceptable'', before conceding that they in fact produce optimal estimates when the missing value is not just missing, but cannot exist, such as the marital quality of an unmarried couple. However, this semantic distinction may not always be clear-cut in practice, and the more pertinent question may be whether missing values are informative. Allison \cite{allison10missing} later acknowledged that missing-indicators may lead to better predictions and their use for that purpose was acceptable. Missing-indicators have also been dismissed in \cite{pigott01review,schafer02missing,graham09missing,aste15techniques}, and are frequently omitted in overviews of missing data strategies \cite{schafer97analysis,enders10applied,eirola14machine,garcia15data,das18handling}.

\subsection{Previous Experiments}
\label{sec_experiments}

Only a handful of experimental comparisons of missing data approaches have included the missing-indicator approach, and these have been limited in scope. \cite{vamplew92missing} and \cite{ng11missing} only use a single dataset with randomly removed values, and base their evaluation on the performance of a single algorithm (respectively a neural network and linear regression). The authors of \cite{pereirabarata19imputation} use three classification algorithms and 22 data\-sets, but again with randomly removed values, explicitly assuming an MCAR context. They conclude that imputation outperforms missing-indicators, but the comparison is not like-for-like, since it involves several forms of imputation but only combines indicator attributes with zero imputation. The authors of \cite{heijdenvander06imputation} compare missing-indicators with zero imputation against several other forms of imputation without missing-indicators on one real dataset, for logistic regression. However, they do not evaluate predictive performance. 

Ding \& Simonoff \cite{ding10investigation} conduct a more extensive investigation, using insights from a series of Monte Carlo simulations to systematically remove values from 36 datasets to simulate different forms of missingness. They use these datasets to compare zero imputation\footnote{Presumably, they use one-hot encoding for categorical attributes, in which case zero imputation is equivalent to treating missing values as a separate category, but they do not state this explicitly.} with indicator attributes against mean/mode imputation without, as well as a number of other missing data approaches, for logistic regression. In addition, the authors evaluate a related representation of missing values\footnote{For categorical values, encoding missing values as a separate category; for numerical values, encoding missing values as an extremely large value that can always be split from the other values.} on the same set of 36 datasets, and on one real-life dataset with missing values, for decision trees. They find that there is strong evidence that representing missing values is the best approach when they are informative; when this is not the case their results show no strong difference.

The comparison by Grzymala-Busse \& Hu \cite{grzymalabusse00comparison} is based on 10 datasets with naturally occurring missing values. However, the setting is purely categorical --- all attributes are transformed into categorical attributes --- the only form of imputation is mode imputation, and the missing value approaches are evaluated on the basis of the LERS classifier (Learning from Examples based on Rough Sets \cite{grzymalabusse88knowledge}).

Marlin \cite{marlin08missing} compares zero imputation with missing-indicators (\e{augmentation with response indicators}) against several forms of imputation without, for logistic regression and neural networks, on the basis of an extensive series of simulations, one dataset with artificially removed values, and three real datasets. For the real datasets, there is no strong difference in performance between the different approaches.

Most recently, building on earlier experiments with simulated regression datasets \cite{josse24consistency,lemorvan21whats}, Perez-Lebel et al. \cite{perezlebel22benchmarking} compare four different imputation techniques with and without missing-indicators (\e{missingness mask}) on seven prediction tasks derived from four real medical datasets, and conclude that missing-indicators consistently improve performance for gradient boosted trees, ridge regression and logistic regression.

We point out that the Missingness in Attribute (MIA) proposal \cite{twala08good} for decision trees and decision tree ensembles can be understood as an implicit combination of missing-indicators with automatic imputation, and has also been shown to outperform imputation without missing-indicators in small-scale experimental studies \cite{josse24consistency,perezlebel22benchmarking}.

Finally, even experimental comparisons of missing data that do not feature the missing-indicator approach generally do not involve more than a handful of real-life datasets with naturally occurring missing values. We have only found the connected works \cite{luengo12choice,luengo12missing}, which feature 21 datasets from the UCI repository, but 12 of these are problematic.\footnote{The target column of the \e{echocardiogram} dataset (`alive-at-1') is supposed to denote whether a patient survived for at least one year, but it doesn't appear to agree with the columns from which it is derived, that denote how long a patient (has) survived and whether they were alive at the end of that period. The \e{audiology} dataset has a large number of small classes with complex labels and should perhaps be analysed with multi-label classification. In addition, it has ordinal attributes where the order of the values is not entirely clear, and three different values that potentially denote missingness (`?', `unmeasured' and `absent'), and it is not completely clear how they relate to each other. The \e{house-votes-84} dataset contains `?' values, but its documentation explicitly states that these values are not unknown, but indicate different forms of abstention. The \e{ozone} dataset is a time-series problem, while the task associated with the \e{sponge} and \e{water-treatment} datasets is clustering, with no obvious target for classification among their respective attributes. Finally, the \e{breast-cancer} (9), \e{cleveland} (7), \e{dermatology} (8), \e{lung-cancer} (5), \e{post-operative} (3) and \e{wisconsin} (16) datasets contain only very few missing values, and any performance difference between missing value approaches on these datasets may to a large extent be coincidental.}


\section{Experimental Setup}
\label{sec_experimental_setup}

To evaluate the effect of the missing-indicator approach on classification performance, we conduct a series of experiments, using the Python machine learning library \e{scikit-learn} \cite{pedregosa11scikitlearn}.

\subsection{Questions}
\label{sec_questions}

The aim of our experiments is to answer the following questions:

\begin{itemize}
 \item Do missing-indicators increase performance, and does it matter which imputation strategy they are paired with?
 \item When do missing-indicators start to become useful in terms of missingness?
 \item Does using mean imputation instead of mode imputation allow for more information to be learned from missing categorical values?
 \item How do missing-indicators compare to model-specific approaches to missing values?
\end{itemize}

\subsection{Evaluation}
\label{sec_evaluation}
We preprocess datasets by standardising numerical attributes and one-hot encoding categorical attributes (as required by the implementations in scikit-learn).

We measure classification performance by performing stratified five-fold cross-validation, repeating this for five different random states (which determine both the dataset splits and the initialisation of algorithms with a random component), and calculating the mean area under the receiver operator curve (AUROC). For multi-class datasets, we use the extension of AUROC defined in \cite{hand01simple}.

To compare two alternatives A and B, we consider the $p$-value of a one-sided Wilcoxon signed-rank test \cite{wilcoxon45individual} on the mean AUROC scores for our selection of datasets. When we compare A vs B, a score below 0.5 means that A increased performance on our selection of datasets; the lower the scores, the more confident we can be that this generalises to other similar datasets. Conversely, a score higher than 0.5 means that A decreased performance on our selection of datasets.

\subsection{Imputation Strategies}
\label{sec_imputation_strategies}

We consider the following three imputation strategies:

\begin{itemize}
 \item \e{Mean/mode imputation} replaces missing values of numerical and categorical attributes by, respectively, the mean and the mode of the non-missing values.
 \item \e{Nearest neighbour imputation} \cite{troyanskaya01missing} replaces missing values of numerical and categorical attributes by, respectively, the mean and the mode of the 5 nearest non-missing values, with distance determined by the corresponding non-missing values for the other attributes.
 \item \e{Iterative imputation}, as implemented in scikit-learn, based on \cite{buuren11mice}, predicts missing values of one attribute on the basis of the other attribute values using a round-robin approach. For numerical attributes, this uses Bayesian ridge regression \cite{tipping01sparse}, initialised with mean imputation, while for categorical attributes, we use logistic regression, initialised with mode imputation.
\end{itemize}

The scikit-learn implementations of nearest neighbour and iterative imputation can currently only impute numerical features, so we had to adapt them for categorical imputation. In all other aspects, we follow the default settings of scikit-learn.\footnote{For the \e{nomao} dataset, iterative imputation diverged, so we had to restrict imputation to the interval $[-100, 100]$.}

\subsection{Classification Algorithms}
\label{sec_classification_algorithms}

We consider the classification algorithms listed in Table~\ref{tab_classifiers}, as implemented in scikit-learn. Hyperparameters take their default values, except for SVM-L, LR and MLP, where we increase the maximum number of iterations to 10\,000 to increase the probability of convergence.

For a number of these algorithms, specific ways have been proposed to handle missing values: e.g. NN-2-D \cite{dixon79pattern}, SVM-G \cite{smieja19generalized}, MLP \cite{tresp94efficient,smieja18processing,ipsen20how} and CART \cite{quinlan89unknown,twala08good}. The purpose of the present experiment is to evaluate the general approach of using imputation with missing-indicators when these solutions have not been implemented, as is the case in scikit-learn.

\begin{table}
\centering
\caption{Classification algorithms.}
\label{tab_classifiers}
\begin{tabular}{p{.15\linewidth}p{.75\linewidth}}
\toprule
Name &     Description \\
\midrule
     NN-1 & Nearest neighbours \cite{fix51discriminatory} with (Boscovich) 1-distance\\
     NN-2 & Nearest neighbours with (Euclidean) 2-distance\\
     NN-1-D & Nearest neighbours with 1-distance, distance-weighted \cite{dudani76distance}\\
     NN-2-D & Nearest neighbours with 2-distance, distance-weighted\\
   SVM-L & Soft-margin Support Vector Machine \cite{cortes95support} with linear kernel\\
   SVM-G & Soft-margin Support Vector Machine with Gaussian kernel \\
        LR & Multinomial logistic regression \cite{cox66some}\\
       MLP & Multilayer perceptron \cite{rosenblatt61principles} with ReLu activation \cite{fukushima69visual}, Glorot initialisation \cite{glorot10understanding} and Adam optimisation \cite{kingma15adam}\\
      CART & Classification and Regression Tree \cite{breiman84classification}\\
        RF & Random Forest \cite{breiman01random}\\
       ERT & Extremely Randomised Trees \cite{geurts06extremely}\\
       ABT & Ada-boosted trees \cite{freund95desiciontheoretic} with SAMME (stagewise additive modeling using a
multi-class exponential loss function) \cite{zhu09multiclass}\\
       GBM & Gradient Boosting Machine \cite{friedman01greedy}\\
\bottomrule
\end{tabular}
\end{table}

\subsection{Datasets}
\label{sec_datasets}

We use twenty real-life datasets with naturally occurring missing values from the UCI repository for machine learning \cite{dua19uci} (Table~\ref{tab_statistics}). These datasets are quite varied --- they cover a number of different domains and contain between 155 and 76\,000 records, between 4 and 590 attributes, between 2 and 21 decision classes and missing value rates between 0.0032 and 0.43.

We have preprocessed these data\-sets in the following manner. We have removed attributes that were non-informative according to the accompanying documentation, as well as identifiers and alternative target values. When it was clear from the description that an attribute was categorical, we have treated it as such, even if it was originally represented with numerals. Conversely, where the possible values of an attribute admitted a semantic order, we have encoded them numerically. We have left binary attributes in their original encoding (categorical or numerical). To enable 5-fold cross-validation, we have removed classes with fewer than 5 records.

\begin{table}
\centering
\caption{Real-life classification datasets with missing values from the UCI repository for machine learning.}
\label{tab_statistics}
\begin{tabular}{lrrrrrllll}
\toprule
            Dataset & Records & Classes & \multicolumn{3}{l}{Attributes} & \multicolumn{3}{l}{Missing value rate} &                          Source \\
                $ $ &     $ $ &     $ $ &        Num & Cat & Total &                Num &    Cat &  Total &                             $ $ \\
\midrule
              adult &   48842 &       2 &          5 &   8 &    13 &                0.0 &  0.017 &  0.010 &          \cite{kohavi96scaling} \\
   agaricus-lepiota &    8124 &       2 &          1 &  21 &    22 &                0.0 &  0.015 &  0.014 &       \cite{schlimmer87concept} \\
        aps-failure &   76000 &       2 &        170 &   0 &   170 &              0.083 &        &  0.083 &       \cite{ferreiracosta16ida} \\
         arrhythmia &     443 &      10 &        279 &   0 &   279 &             0.0032 &        & 0.0032 &      \cite{guvenir97supervised} \\
              bands &     540 &       2 &         19 &  15 &    34 &              0.054 &  0.054 &  0.054 &        \cite{evans94overcoming} \\
                ckd &     400 &       2 &         14 &  10 &    24 &               0.14 &  0.059 &   0.11 &       \cite{rubini15generating} \\
                crx &     690 &       2 &          6 &   9 &    15 &             0.0060 & 0.0068 & 0.0065 &     \cite{quinlan87simplifying} \\
        dress-sales &     500 &       2 &          3 &   9 &    12 &               0.20 &   0.19 &   0.19 &                                 \\
            exasens &     399 &       4 &          7 &   0 &     7 &               0.43 &        &   0.43 &   \cite{soltanizarrin20invitro} \\
                hcc &     165 &       2 &         49 &   0 &    49 &               0.10 &        &   0.10 &              \cite{santos15new} \\
      heart-disease &    1611 &       2 &         13 &   1 &    14 &               0.18 &    0.0 &   0.17 &   \cite{detrano89international} \\
          hepatitis &     155 &       2 &         19 &   0 &    19 &              0.057 &        &  0.057 &       \cite{efron81statistical} \\
        horse-colic &     368 &       2 &         19 &   1 &    20 &               0.25 &   0.39 &   0.26 &       \cite{mcleish90enhancing} \\
mammographic-masses &     961 &       2 &          2 &   2 &     4 &              0.042 &  0.041 &  0.042 &        \cite{elter07prediction} \\
                 mi &    1700 &       8 &        111 &   0 &   111 &              0.085 &        &  0.085 & \cite{golovenkin20trajectories} \\
              nomao &   34465 &       2 &         89 &  29 &   118 &               0.38 &   0.37 &   0.38 &       \cite{candillier12design} \\
      primary-tumor &     330 &      15 &         16 &   1 &    17 &              0.029 &   0.20 &  0.039 &       \cite{cestnik87assistant} \\
              secom &    1567 &       2 &        590 &   0 &   590 &              0.045 &        &  0.045 &        \cite{mccann08causality} \\
            soybean &     683 &      19 &         22 &  13 &    35 &              0.099 &  0.096 &  0.098 &      \cite{michalski80learning} \\
        thyroid0387 &    9172 &      18 &          7 &  16 &    23 &               0.22 & 0.0021 &  0.069 &       \cite{quinlan86inductive} \\
\bottomrule
\end{tabular}
\end{table}

\begin{table}
\centering
\caption{One-sided $p$-values, imputation with missing-indicators versus without.}
\label{tab_p_values_indicators}
\begin{tabular}{llll}
\toprule
Classifier & \multicolumn{3}{l}{Imputation strategy} \\
       $ $ &           Mean/mode & Neighbours & Iterative \\
\midrule
      NN-1 &              0.0088 &     0.0015 &    0.0017 \\
      NN-2 &               0.015 &     0.0024 &   0.00048 \\
    NN-1-D &              0.0045 &     0.0019 &    0.0011 \\
    NN-2-D &              0.0019 &     0.0031 &   0.00027 \\
     SVM-L &                0.13 &       0.27 &     0.099 \\
     SVM-G &              0.0032 &     0.0027 &    0.0021 \\
        LR &               0.079 &      0.063 &     0.068 \\
       MLP &              0.0027 &     0.0063 &    0.0056 \\
      CART &                0.44 &       0.39 &      0.40 \\
        RF &               0.038 &      0.051 &      0.17 \\
       ERT &                0.28 &     0.0099 &     0.026 \\
       ABT &               0.089 &      0.078 &      0.47 \\
       GBM &                0.17 &      0.012 &      0.36 \\
\bottomrule
\end{tabular}
\end{table}

\section{Results and Discussion}
\label{sec_results}

Using the experimental setup detailed in the previous section, we now try to answer the questions listed in Subsection~\ref{sec_questions}.

\subsection{Do Missing-Indicators Increase Performance, and Does It Matter Which Imputation Strategy They Are Paired With?}
\label{sec_results_indicators}

The $p$-values obtained by comparing imputation with and without missing-indicators are displayed in Table~\ref{tab_p_values_indicators}. Missing-indicators generally lead to increased performance --- with the notable exception of CART, to which we return below. The more complicated imputation strategies do not result in much better results than mean/mode imputation when we pair imputation with missing-indicators (Table~\ref{tab_p_values_imputation}). At best, nearest neighbour and iterative imputation only lead to a modest improvement, and for many classifiers, they actually decrease performance. Therefore, we focus on mean/mode imputation for the remainder of this section.

\begin{table}
\centering
\caption{One-sided $p$-values, missing-indicators with iterative and nearest neighbour versus mean/mode imputation.}
\label{tab_p_values_imputation}
\begin{tabular}{lll}
\toprule
Classifier & \multicolumn{2}{l}{Imputation strategy} \\
       $ $ &          Neighbours & Iterative \\
\midrule
      NN-1 &                0.94 &      0.15 \\
      NN-2 &                0.78 &      0.19 \\
    NN-1-D &                0.97 &      0.55 \\
    NN-2-D &                0.84 &      0.23 \\
     SVM-L &                0.53 &      0.61 \\
     SVM-G &                0.47 &      0.94 \\
        LR &                0.40 &      0.83 \\
       MLP &                0.30 &      0.55 \\
      CART &                0.69 &      0.79 \\
        RF &                0.61 &      0.86 \\
       ERT &                0.61 &      0.64 \\
       ABT &                0.33 &      0.78 \\
       GBM &                0.93 &      0.85 \\
\bottomrule
\end{tabular}
\end{table}

\begin{figure}
\centering
\subfloat[adult]{
\includegraphics[width=.45\linewidth]{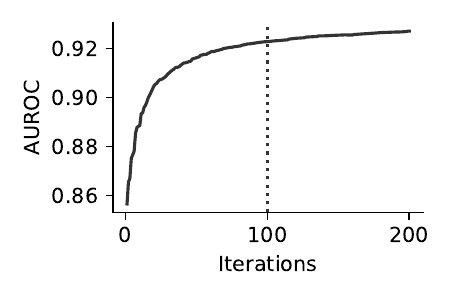}
\label{fig_gbm_underfitting}}%
\hfil
\subfloat[mammographic-masses]{
\includegraphics[width=.45\linewidth]{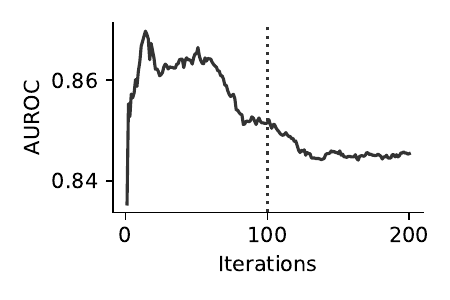}
\label{fig_gbm_overfitting}}%
\caption{GBM test AUROC for two illustrative datasets, using mean/mode imputation without missing-indicators, for one random state and one cross-validation fold. The default hyperparameter value of 100 iterations leads to under- \subref{fig_gbm_underfitting} and overfitting \subref{fig_gbm_overfitting}.}
\label{fig_gbm_auroc}
\end{figure}

A possible reason for the failure of missing-indicators to increase performance with CART, is that by default, the scikit-learn implementation of this classifier does not perform pruning, making it prone to overfitting. To test this hypothesis, we repeat our experiment for CART and mean imputation, but this time we apply cost complexity pruning ($\alpha = 0.01$). This clearly improves performance ($p = 0.0069$ without missing-indicators, $p = 0.015$ with missing-indicators), and now missing-indicators have a slight advantage ($p = 0.23$).

We have also taken a closer look at ERT and GBM, for which the performance increase from missing-indicators is not very significant. For ERT, this may be due to underfitting. If we increase the number of trees from the default 100 to 1000, this improves performance ($p = 0.0011$ without missing-indicators, $p = 0.0032$ with missing-indicators), and makes the advantage of missing-indicators somewhat clearer ($p = 0.092$).

For GBM, the default choice of 100 iterations of gradient descent can lead to both under- or overfitting, depending on the dataset (Fig.~\ref{fig_gbm_auroc}). We believe that it is generally preferable to continue training until an early-stopping criterion is met. However, applying the same criterion as with MLP\footnote{Setting aside 10\% of the data for validation, stopping when validation loss has not decreased by at least 0.0001 for ten iterations, with a maximum of 10\,000 iterations.} does not improve performance over the default of 100 ($p = 0.81$ without missing-indicators, $p = 0.85$ with missing-indicators) and does not change the relative advantage due to missing-indicators ($p = 0.20$).

\subsection{When Do Missing-Indicators Start to Become Useful in Terms of Missingness?}
\label{sec_missing_thresholds}

The theoretical motivation for representing missing values through missing-indicators is that this allows classifiers to learn the information encoded in their distribution. In principle, this should be easier when there are more examples to learn from. We can use this principle to obtain a better understanding of when missing-indicators might be useful on a per-attribute level.

The challenge that we have to overcome is that we would like to study individual attributes, but classification performance is measured on the dataset level. We tackle this by studying datasets with only one attribute with missing values, allowing us to investigate the relation between the properties of the attribute and classification performance on the dataset.

We conduct the following experiment. For each attribute with missing values in each dataset, we reduce the original dataset by removing all other attributes with missing values. We thus obtain 1148 reduced datasets with only one attribute with missing values, onto which we apply each of our classifiers (with pruning for CART, 1000 trees for ERT and early-stopping for GBM) and consider whether missing-indicators increase or decrease AUROC (we dismiss ties). Finally, for each classifier we fit a logistic regression model with cluster robust covariance (clustered by the originating dataset), with the following potential parameters: categoricalness (whether the attribute is categorical) and either the number of missing values (log-transformed) or the missing rate. We use the Akaike information criterion \cite{akaike73information} to decide whether to select these parameters.

We find that for most classifiers, either the absolute or the relative number of missing values is an informative parameter with positive coefficient. For MLP, neither parameter is informative, while for RF, the number of missing values is an informative parameter with negative coefficient, for which we have no explanation at present. For every classifier, categoricalness is an informative parameter with positive coefficient, meaning that missing-indicators are more beneficial for categorical than for numerical attributes.

The fitted logistic regression models allow us to calculate attribute-specific thresholds above which missing-indicators are more likely than not to increase AUROC, for all classifiers except MLP and RF (Table~\ref{tab_missing_thresholds}). In many cases, these thresholds are 1 or 0.0, indicating that missing-indicators are always likely to increase AUROC.

\begin{table}
\centering
\caption{Thresholds above which missing-indicators are more likely than not to increase AUROC, in terms of the absolute number of missing values or the missing rate.}
\label{tab_missing_thresholds}
\begin{tabular}{lrrrr}
\toprule
Classifier & \multicolumn{2}{l}{Missing values} & \multicolumn{2}{l}{Missing rate} \\
           &            Cat &   Num &          Cat &  Num \\
\midrule
      NN-1 &              1 &   302 &              &      \\
      NN-2 &              2 &   130 &              &      \\
    NN-1-D &              1 &   291 &              &      \\
    NN-2-D &              1 &    73 &              &      \\
     SVM-L &                &       &          0.0 &  0.0 \\
     SVM-G &                &       &          0.0 & 0.40 \\
        LR &                &       &          0.0 &  0.0 \\
      CART &                &       &          0.0 & 0.12 \\
       ERT &                &       &          0.0 &  1.0 \\
       ABT &              1 & 23200 &              &      \\
       GBM &                &       &          0.0 &  0.0 \\
\bottomrule
\end{tabular}
\end{table}

\subsection{Does Using Mean Imputation Instead of Mode Imputation Allow for More Information to Be Learned from Missing Categorical Values?}
\label{sec_numerical_imputation_categorical_values}

As indicated above, missing-indicators are generally more likely to increase performance for categorical than for numerical attributes. A potential explanation for this is the fact that the mode of a categorical attribute is one of the non-missing values, whereas the mean of a numerical attribute is generally not equal to one of the non-missing values. Therefore, categorical imputation renders missing values truly indistinguishable from non-missing values, whereas numerical imputation does not --- the information expressed by missing values may be partially recoverable, as argued by Le Morvan et al. \cite{lemorvan21whats} and discussed in the Introduction.

We can achieve a similar partial representation of missing categorical values by changing the order in which we perform imputation and one-hot encoding, i.e. by performing numerical imputation on one-hot encoded categorical attributes with missing values. For imputation without missing-indicators, this indeed leads to better performance for some classifiers, while in combination with missing-indicators, it does not make much of a difference (Table~\ref{tab_p_values_categorical})\footnote{LR is an exception here, we have no explanation for this.}.

\begin{table}
\centering
\caption{One-sided $p$-values, mean imputation after one-hot encoding versus mode imputation of missing categorical values.}
\label{tab_p_values_categorical}
\begin{tabular}{lllllllllllll}
\toprule
Classifier &  Without --- &  With missing-indicators \\
\midrule
      NN-1 &        0.020 &                    0.077 \\
      NN-2 &         0.14 &                    0.031 \\
    NN-1-D &        0.016 &                     0.12 \\
    NN-2-D &         0.16 &                    0.031 \\
     SVM-L &         0.43 &                     0.57 \\
     SVM-G &         0.17 &                     0.56 \\
        LR &         0.81 &                    0.057 \\
       MLP &         0.16 &                     0.60 \\
      CART &         0.44 &                     0.30 \\
        RF &        0.046 &                     0.57 \\
       ERT &        0.030 &                     0.95 \\
       ABT &         0.48 &                     0.62 \\
       GBM &        0.077 &                     0.54 \\
\bottomrule
\end{tabular}
\end{table}

\subsection{How Do Missing-Indicators Compare to Model-Specific Approaches to Missing Values?}
\label{sec_native}

While not the primary focus of this paper, we may also wonder how the missing-indicator approach compares to model-specific approaches to missing values. For CART and RF, we consider the proposal by \cite{twala08good}, that a decision tree should evaluate two variants of each split, with missing values sent to either side. This has been implemented in the latest version of scikit-learn (1.4.0), which was released after the previous experiments in this section were conducted. In addition, we have modified the implementation of the nearest neighbour classifier in scikit-learn to obtain the approach labelled as `normal' in \cite{dixon79pattern}. This calculates the distance between two records by linearly extrapolating the distance calculated only on the basis of all non-missing feature values. We note that every model-specific approach is different --- we expect that their effect on classification performance will differ from case to case --- so our evaluation of these two approaches only serves an illustrative purpose.

We find (Table~\ref{tab_p_values_native}, Test 1) that the model-specific approach for the nearest neighbour classifiers performs significantly worse than mean/mean imputation with missing-indicators. In contrast, there is no difference for CART, and the model-specific approach appears to perform better for RF. We can also ask whether these model-specific approaches benefit from adding missing-indicators --- here this only appears to be the case for the nearest neighbour classifiers (Table~\ref{tab_p_values_native}, Test 2), i.e. when the model-specific approach performs badly. However, even with missing-indicators the model-specific approach for the nearest neighbour classifiers does not perform better than mean/mean imputation with missing indicators (Table~\ref{tab_p_values_native}, Test 3).

\begin{table}
\centering
\caption{One-sided $p$-values, model-specific missing value approaches. Test 1: Mean/mean imputation with missing-indicators vs model-specific approach without; Test 2: Model-specific approach with missing-indicators vs model-specific approach without; Test 3: Model-specific approach vs mean/mean imputation, both with missing-indicators.}
\label{tab_p_values_native}
\begin{tabular}{llll}
\toprule
Classifier & Test 1 & Test 2 & Test 3 \\
\midrule
NN-1 & 0.00036 & 0.00017 & 0.56 \\
NN-2 & 0.00074 & 0.00015 & 0.94 \\
NN-1-D & 0.00020 & 0.00015 & 0.89 \\
NN-2-D & 0.00023 & 0.00011 & 0.93 \\
CART & 0.50 & 0.86 & 0.66 \\
RF & 0.92 & 0.60 & 0.092 \\
\bottomrule
\end{tabular}
\end{table}

\section{Conclusion}
\label{sec_conclusion}

We have presented the first large-scale experimental evaluation of the effect of the missing-indicator approach on classification performance, conducted on real datasets with naturally occurring missing values, paired with three different imputation techniques. The central question was whether, on balance, more benefit can be derived from the additional information encoded in a representation of missing values, or from the lower-dimensional projection of the data obtained by omitting missing-indicators.

On the whole, missing-indicators increase performance for the classification algorithms that we considered. An exception was CART, which suffers from overfitting in its default scikit-learn configuration. When pruning is applied, missing-indicators do increase performance. For ERT, the advantage of missing-indicators becomes more significant when underfitting is controlled.

We also found that, in the presence of missing-indi\-cators, nearest neighbour and iterative imputation do not significantly increase performance over simple mean/mode imputation. This is a useful finding, because implementations of more sophisticated imputation strategies may not always be available to practitioners working in different frameworks, or easy to apply.

In a follow-up experiment, we determined attribute-specific missingness thresh\-olds, above which missing-indi\-cators are more likely than not to increase performance. For categorical attributes, this threshold is generally very low, while for numerical attributes, there is more variation among classifiers, in particular as to whether this threshold is absolute or relative to the total number of records.

The greater usefulness of missing-indicators for categorical than for numerical attributes can be explained by the fact that the mean of a numerical attribute is not generally identical to any of the non-missing values, and that mean imputation therefore preserves some of the information of missing values. This is supported by the results of a further experiment, which showed that, in the absence of missing-indicators, applying mean imputation to one-hot encoded categorical attributes results in somewhat better performance than mode imputation.

While we have mainly considered the use of missing-indicators with imputation, there also exist model-specific solutions for missing values, that can in turn be combined with missing-indicators. Whether missing-indicators outperform these model-specific approaches has to be determined on a case-by-case basis. This was illustrated by our third follow-up experiment for nearest neighbour and decision tree classifiers.

We conclude that the combination of mean/mode imputation with missing-indicators is a safe default approach towards missing values in classification tasks. While over- or underfitting is a concern for certain classifiers, it is a concern for these classifiers with or without missing-indicators. However, practitioners may want to omit miss\-ing-indicators when the classification algorithm to be used has a special provision for missing values, when the missingness thresholds that we determined are not met, or on the basis of specific information about the distribution of missing values in the dataset. The use of missing-indicators can also be combined with dimensionality reduction algorithms to increase the information density of the resulting dataset.

The problem of missing data has been the subject of a rich body of theoretical literature. We hope to have contributed with this paper to the practical evaluation of some of that theory. In particular, we are happy to have identified twenty real-life datasets with missing values, and hope that in the future, more such datasets will be collected.

\begin{credits}

\subsubsection{Data and code.} Datasets and the code to reproduce our experiments are available at \url{https://cwi.ugent.be/~oulenz/code/lenz-2024-no.tar.gz}.

\subsubsection{\ackname} The research reported in this paper was conducted with the financial support of the Odysseus programme of the Research Foundation -- Flanders (FWO). This publication is part of the project Digital Twin with project number P18-03 of the research programme TTW Perspective, which is (partly) financed by the Dutch Research Council (NWO). We would like to express our thanks to Geert van der Heijden for answering a question about \cite{heijdenvander06imputation}.

\subsubsection{\discintname}
The authors have no competing interests to declare that are relevant to the content of this article

\end{credits}

\DeclareRobustCommand{\VAN}[3]{#3}

\bibliographystyle{splncs04}
\bibliography{20230215_missing-indicators}

\end{document}